# Learning the Structure of Dynamic Probabilistic Networks

**Nir Friedman**    **Kevin Murphy**    **Stuart Russell**
Computer Science Division, U. of California, Berkeley, CA 94720
{nir,murphyk,russell}@cs.berkeley.edu

## Abstract

*Dynamic probabilistic networks* are a compact representation of complex stochastic processes. In this paper we examine how to learn the *structure* of a DPN from data. We extend structure scoring rules for standard probabilistic networks to the dynamic case, and show how to search for structure when some of the variables are hidden. Finally, we examine two applications where such a technology might be useful: predicting and classifying dynamic behaviors, and learning causal orderings in biological processes. We provide empirical results that demonstrate the applicability of our methods in both domains.

## 1 INTRODUCTION

*Probabilistic networks* (PNs), also known as *Bayesian networks* or *belief networks*, are already well-established as representations of domains involving uncertain relations among several random variables. Somewhat less well-established, but perhaps of equal importance, are *dynamic probabilistic networks* (DPNs), which model the stochastic evolution of a set of random variables over time [5]. DPNs have significant advantages over competing representations such as Kalman filters, which handle only unimodal posterior distributions and linear models, and hidden Markov models (HMMs), whose parameterization grows exponentially with the number of state variables. For example, [31] show that DPNs can outperform HMMs on standard speech recognition tasks.

PNs and DPNs are defined by a graphical structure and a set of parameters, which together specify a joint distribution over the random variables. Algorithms for learning the parameters of PNs [1, 21] and DPNs [1, 14] are becoming widely used. These algorithms typically use either gradient methods or EM, and can handle hidden variables and missing values.

Algorithms for learning the graphical structure, on the other hand, have until recently been restricted to networks with *complete data*, i.e., where the values of all variables are specified in each training case [4, 15]. Friedman [10, 11] has developed the *Structural EM* (SEM) algorithm for learning PN structure from data with hidden variables and missing values. SEM combines structural and parametric modification within a single EM process, and appears to be substantially more effective than previous approaches based on parametric EM operating within an outer-loop structural search. SEM can be shown to find local optima defined by a scoring function that combines the likelihood of the data with a structural penalty that discourages overly complex networks. This property holds for both the *Bayesian Information Criterion* (BIC) score [28], a variant of Minimum Description Length (MDL) scoring, and the BDe score [15], a Bayesian metric that uses an explicit prior over networks.

In this paper, we extend the BIC and BDe scores to handle the problem of learning DPN structure from complete data. More importantly, we extend the SEM algorithm to learn DPNs from incomplete data with both scores. Given partial observations of a set of random variables over time, the algorithm constructs a DPN (possibly including additional hidden variables) that fits the observations as well as possible. The addition of hidden variables in DPNs is particularly important as many processes—human decision making, speech generation, disease processes, for example—are only partially observable.

We begin with formal definitions of PNs and DPNs. Section 3 discusses the complete-data case, developing scores for DPNs so that existing algorithms for learning PN structures can be applied directly. In Section 4, we handle the case of incomplete data and show how to extend SEM to DPNs. Computation of the necessary sufficient statistics is highlighted as a principal bottleneck. Section 5 describes two applications: learning simple models of human driving behavior from videotapes, and learning models of biological processes from very sparse observations.

## 2 PRELIMINARIES

We start with a short review of probabilistic networks and dynamic probabilistic networks.

We will be concerned with distributions over sets of discrete random variables, where each variable $X_i$ may take on values from a finite set, denoted by $Val(X_i)$. We denote the size of $Val(X_i)$ by $||X_i||$. We use capital letters, such as $X, Y, Z$, for variable names and lowercase letters $x, y, z$ to denote specific values taken by those variables. Sets of variables are denoted by boldface capital letters $\mathbf{X}, \mathbf{Y}, \mathbf{Z}$, with sets of values denoted by boldface lowercase letters $\mathbf{x}, \mathbf{y}, \mathbf{z}$. For a given network, we will use $\mathbf{X} = \{X_1, \ldots, X_n\}$ to denote the variables of the network in topological order, and $P$ to denote a joint probability distribution over the variables

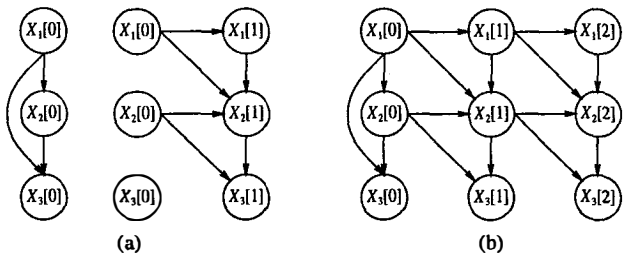

Figure 1: (a) A prior network and transition network defining a DPN for the attributes $X_1$, $X_2$, $X_3$. (b) The corresponding "unrolled" network.

in **X**.

A *probabilistic network* (PN) is an annotated directed acyclic graph that encodes a joint probability distribution over **X**. Formally, a PN for **X** is a pair $B = (G, \Theta)$. The first component, $G$, is a directed acyclic graph whose vertices correspond to the random variables $X_1, \ldots, X_n$ that encodes the following set of conditional independence assumptions: each variable $X_i$ is independent of its non-descendants given its parents $\mathbf{Pa}(X_i)$ in $G$. The second component, $\Theta$, represents the set of parameters that quantifies the network. In the simplest case, it contains a parameter $\theta_{i,j_i,k_i} = \Pr(X_i = k_i | \mathbf{Pa}(X_i) = j_i)$ for each possible value $k_i$ of $X_i$ and each possible set of values $j_i$ of $\mathbf{Pa}(X_i)$. Each conditional probability distribution can be represented as a table, called a CPT (conditional probability table). Representations which require fewer parameters, such as noisy-ORs [27] or trees [12], are also possible — indeed, we use them in the experimental results section — but, for simplicity of notation, we shall stick to the CPT case.

Given $G$ and $\Theta$, a PN $B$ defines a unique joint probability distribution over **X** given by:

$$P_B(x_1, \ldots, x_n) = \prod_{i=1}^{n} P_B(x_i | \mathbf{pa}(X_i))$$

A PN describes a probability distribution over a fixed set of variables. DPNs extend this representation to model temporal processes. For simplicity, we assume that changes occur between discrete time points that are indexed by the non-negative integers. We assume that $\mathbf{X} = \{X_1, \ldots, X_n\}$ is a set of attributes that the process changes. $X_i[t]$ is the random variable that denotes the value of the attribute $X_i$ at time $t$, and $\mathbf{X}[t]$ is the set of random variables $X_i[t]$.

To represent beliefs about the possible trajectories of the process, we need a probability distribution over the random variables $\mathbf{X}[0] \cup \mathbf{X}[1] \cup \mathbf{X}[2] \cup \ldots$. Of course, such a distribution can be extremely complex. In this paper, we assume the process is *Markovian* in **X**, i.e., $P(\mathbf{X}[t+1] | \mathbf{X}[0], \ldots, \mathbf{X}[t]) = P(\mathbf{X}[t+1] | \mathbf{X}[t])$. We also assume that the process is *stationary*, i.e., that the *transition probability* $P(\mathbf{X}[t+1] | \mathbf{X}[t])$ is independent of $t$.

Given these assumptions, a DPN that represents the joint distribution over all possible trajectories of a process consists of two parts:

- a prior network $B_0$ that specifies a distribution over initial states $\mathbf{X}[0]$; and

- a transition network $B_\rightarrow$ over the variables $\mathbf{X}[0] \cup$ $\mathbf{X}[1]$ that is taken to specify the transition probability $P(\mathbf{X}[t+1] | \mathbf{X}[t])$ for all $t$.

Figure 1(a) shows a simple example.[1] In the transition network (but not the prior network), the variables in $\mathbf{X}[0]$ have no parents. The transition probability implied by such a network is:

$$P_{B_\rightarrow}(\mathbf{x}[1] | \mathbf{x}[0]) = \prod_{i=1}^{n} P_{B_\rightarrow}(x_i[1] | \mathbf{pa}(X_i[1])).$$

A DPN defined by a pair $(B_0, B_\rightarrow)$ corresponds to a semi-infinite network over the variables $\mathbf{X}[0], \ldots, \mathbf{X}[\infty]$. In practice, we reason only about a finite interval $0, \ldots, T$. To do this, we can notionally "unroll" the DPN structure into a PN over $\mathbf{X}[0], \ldots, \mathbf{X}[T]$. In slice 0, the parents of $X_i[0]$ are those specified in the prior network $B_0$; in slice $t + 1$, the parents of $X_i[t+1]$ are those nodes in slices $t$ and $t+1$ corresponding to the parents of $X_i[1]$ in $B_\rightarrow$. We copy the conditional distributions for these variables in a similar manner. Figure 1(b) shows the result of unrolling the network in Figure 1(a) for 3 time slices. Given a DPN model, the joint distribution over $\mathbf{X}[0], \ldots, \mathbf{X}[T]$ is

$$P_B(\mathbf{x}[0], \ldots, \mathbf{x}[T]) = P_{B_0}(\mathbf{x}[0]) \prod_{t=0}^{T-1} P_{B_\rightarrow}(\mathbf{x}[t+1] | \mathbf{x}[t]) \quad (1)$$

where $P_{B_\rightarrow}(\mathbf{x}[t+1] | \mathbf{x}[t])$ is obtained in the obvious way from the transition model.

## 3 LEARNING from Complete Data

In this section we develop the theory for learning DPNs from *complete data*. We begin with a brief review of how one learns standard PNs. Then we derive the details of the BIC and BDe score for DPNs and finally we discuss how these are optimized in the search for good structures. The upshot of this section is that learning DPNs from complete data uses the same techniques as learning PNs from complete data. Learning DPNs is not quite the same as applying PN methods to the unrolled network, however, due to the constraint of repeated structure and repeated parameters.

### 3.1 LEARNING PNS

The problem of learning a probabilistic network is stated as follows. Given a *training set* $D$ of instances of **X**, find a network $B = (G, \Theta)$ that *best matches* $D$. The notion of "best match" is defined using a scoring function. Several different scoring functions have been proposed in the literature. The most frequently used are the *Bayesian Information Criterion* [28] and the BDe score [15]. Both of these scores combine the *likelihood* of the data according to the network, $L(B : D) = \log \Pr(D \mid B)$, with some penalty relating to the complexity of the network. When learning the structure of PNs, a complexity penalty is essential since the maximum-likelihood network is usually the completely connected network.

---

[1]Some authors define DPNs using just the transition network, assuming that all "slices," including slice 0, have the same structure. In general, however, the prior distribution on $\mathbf{X}[0]$ can have a quite different independence structure from other slices, since it may represent either the way in which the process is initialized or the dependencies induced among the variables in $\mathbf{X}[0]$ by the unobserved portion of the process before $t = 0$. For example, if $t = 0$ represents an arbitrary starting point taken from an infinite process, then it is reasonable to use a dependency structure for $\mathbf{X}[0]$ that reflects the stationary distribution of the Markov chain defined by the transition network. One can combine the prior and transition networks into a single two-slice network, but, because the learning processes for the prior and transition models are distinct, we have chosen to represent them separately.

The BIC and BDe scores are derived from the posterior probability of the network structure. Let the random variable $G$ range over the possible network structures that might in fact obtain in the real world. Then, using Bayes' rule, the posterior distribution over $G$ is

$$\Pr(G \mid D) \propto \Pr(D \mid G) \Pr(G)$$

where the likelihood of the data given a network structure can be computed by conditioning on the associated network parameters:

$$\Pr(D \mid G) = \int \Pr(D \mid G, \Theta) \Pr(\Theta \mid G) d\Theta. \quad (2)$$

Obviously, specifying the parameter priors and evaluating this integral can be difficult.

One approach to avoiding full computation of the integral in (2) is to examine the asymptotic behavior of this term. Given a large number of data points, the posterior probability is insensitive to the choice of prior (assuming that the prior does not give probability zero to any event). Schwarz [28] derives the following asymptotic estimate for well-behaved priors:

$$\log \Pr(D \mid G) = \log \Pr(D \mid G, \hat{\Theta}_G) - \frac{\log N}{2} \#G + O(1), \quad (3)$$

where $\hat{\Theta}_G$ are the parameter settings for $G$ that maximize the likelihood of the data, $N$ is the number of training instances, $\#G$ is the *dimension* of $G$ (which in the case of complete data is just the number of parameters), and $O(1)$ is a constant term which is independent of $N$ and $G$. The BIC score uses Equation (3) to rank candidate network structures. Notice that it obviates the need for parameter priors, and the prior on structures is reduced to counting parameters.[2]

An alternative approach is to evaluate (2) in closed form given a restricted family of priors [3, 4, 15]. Roughly speaking, the prior on the parameters for a given structure $G$ is assumed to factor into independent priors over the parameters for each conditional distribution $P(X_i \mid \mathbf{pa}(X_i))$. When the data is complete, this implies that the posterior factors in the same way. Thus, each conditional distribution can be updated and scored separately. The score can be computed in closed form if we assume in addition that the prior for each conditional distribution is from a conjugate family, such as the Dirichlet distribution. The details appear below.

Since we might examine a large number of possible network structures, we would like to avoid having to assign a prior distribution over parameters for each possible structure. [15] provide a set of assumptions that allow the parameter prior for all structures to be specified using a single network with Dirichlet priors, together with a single "virtual data count" that describes the confidence in that prior.

---

[2] A similar formula arises from the *Minimum Description Length* (MDL) principle [20].

This approach has the desirable property that the scores of two networks that are equivalent (i.e., describe the same set of independence assumptions) are the same. Finally, to compute the full BDe score, a simple description-length penalty corresponding to $\Pr(G)$ is added in.

## 3.2  BIC SCORE FOR DPNS

We now describe the BIC score for DPNs. Unsurprisingly, the results here mirror the results for PNs.

Throughout this section, we assume that we are given a training set $D$ consisting of $N_{\text{seq}}$ complete observation sequences. The $\ell$th such sequence has length $N_\ell$ and specifies values for the variables $\mathbf{x}^\ell[0], \ldots, \mathbf{x}^\ell[N_\ell]$. Such a dataset give us $N_{\text{seq}}$ instances of initial slices, from which we can train $B_0$, and $N = \sum_\ell N_\ell$ instances of transitions, from which we can train $B_\to$.

We start by introducing some notation. Let us define

$$\theta^{(0)}_{i,j'_i,k'_i} = \Pr(X_i[0] = k'_i \mid \mathbf{Pa}(X_i[0]) = j'_i)$$

and similarly

$$\theta^\to_{i,j_i,k_i} = \Pr(X_t[t] = k_i \mid \mathbf{Pa}(X_i[t]) = j_i)$$

for $t = 1, \ldots, T$. We will also need some notation for the sufficient statistics for each family,

$$N^{(0)}_{i,j'_i,k'_i} = \sum_\ell I(X_i[0] = k'_i, \mathbf{Pa}(X_i[0]) = j'_i; \mathbf{x}^\ell)$$

and

$$N^\to_{i,j_i,k_i} = \sum_\ell \sum_t I(X_i[t] = k_i, \mathbf{Pa}(X_i[t]) = j_i; \mathbf{x}^\ell)$$

where $I(\cdot; \mathbf{x}^\ell)$ is an indicator function which takes on value 1 if the event $\cdot$ occurs in sequence $\mathbf{x}^\ell$, and 0 otherwise.

Using (1), and rearranging terms, we find that the likelihood function *decomposes* according to the structure of the DPN, just as with PNs:

$$\Pr(D \mid G, \Theta_G) = \prod_i \prod_{j'_i} \prod_{k'_i} \left(\theta^{(0)}_{i,j'_i,k'_i}\right)^{N^{(0)}_{i,j'_i,k'_i}} \times$$
$$\prod_i \prod_{j_i} \prod_{k_i} \left(\theta^\to_{i,j_i,k_i}\right)^{N^\to_{i,j_i,k_i}} \quad (4)$$

and hence the log-likelihood is given by

$$L(B : D) = \sum_i \sum_{j'_i} \sum_{k'_i} N^{(0)}_{i,j'_i,k'_i} \log \theta^{(0)}_{i,j'_i,k'_i} +$$
$$\sum_i \sum_{j_i} \sum_{k_i} N^\to_{i,j_i,k_i} \log \theta^\to_{i,j_i,k_i} \quad (5)$$

This decomposition facilitates the computation of the BIC and BDe scores in several ways. First, note that the likelihood is expressed as a sum of terms, where each term depends only on the conditional probability of a variable given a particular assignment to its parents. Thus, if we want to find the maximum likelihood parameters, we can maximize within each family independently. Second, the decomposition implies that we can learn $B_0$ independently of $B_\to$. Finally, we can learn $B_\to$ in exactly the same manner as learning a PN for a set of samples of transitions.

We now make these arguments more precise. Using the standard maximum likelihood estimate for multinomial distributions, we immediately get the following expression for $\hat{\Theta}_G$.

$$\hat{\theta}^{(0)}_{i,j'_i,k'_i} = \frac{N^{(0)}_{i,j'_i,k'_i}}{\sum_{k'_i} N^{(0)}_{i,j'_i,k'_i}}$$

and similarly for the transition case.

In the case of CPTs, the number of parameters in the network is given by

$$\#G = \#G_0 + \#G_\rightarrow$$

where

$$\#G_0 = \sum_i \sum_{p \in \mathbf{Pa}(X_i[0])} \|X_p\| \times (\|X_i[0]\|) - 1)$$

and similarly for the transition case.

Finally, substituting into (3), we find that the BIC score is given by

$$BIC(G : D) = BIC_0 + BIC_\rightarrow \qquad (6)$$

where

$$BIC_0 = \sum_i \sum_{j'_i} \sum_{k'_i} N^{(0)}_{i,j'_i,k'_i} \log \hat{\theta}^{(0)}_{i,j'_i,k'_i} - \frac{\log N_{seq}}{2} \#G_0$$

and

$$BIC_\rightarrow = \sum_i \sum_{j_i} \sum_{k_i} N^{\rightarrow}_{i,j_i,k_i} \log \hat{\theta}^{\rightarrow}_{i,j_i,k_i} - \frac{\log N}{2} \#G_\rightarrow$$

Note that the penalty for families in the original time slice is a function of the number of sequences $N_{seq}$, since we only observe that many examples for this part of the model; whereas the penalty for the transition model is a function of $N$, the total number of transitions observed.

### 3.3  BDe SCORE FOR DPNS

Recall that to compute the BDe score, we need to evaluate the following integral.

$$\Pr(D \mid G) = \int \Pr(D \mid G, \Theta_G) \Pr(\Theta_G \mid G) d\Theta_G$$

The first term inside the integral decomposes as in Equation 4. If we assume that the prior over each conditional probability is independent of the rest, then the prior term also decomposes as

$$\Pr(\Theta_G \mid G) = \prod_i \prod_{j'_i} \Pr(\theta^{(0)}_{i,j'_i,k'_i}) \times \prod_i \prod_{j_i} \Pr(\theta^{\rightarrow}_{i,j_i,k_i})$$

Inserting this into the preceding equation, we see that the entire expression consists of an integral over the product of independent terms. Hence $\Pr(D \mid G)$ can be written as a product of two integrals,

$$\prod_i \prod_{j'_i} \int \prod_{k'_i} \left(\theta^{(0)}_{i,j'_i,k'_i}\right)^{N^{(0)}_{i,j'_i,k'_i}} \times \Pr(\theta^{(0)}_{i,j'_i,k'_i}) \times d\theta^{(0)}_{i,j'_i,k'_i}$$

and similarly for the transition case.

In order to obtain a closed-form solution, we assume *Dirichlet* priors [6]. A Dirichlet prior for a multinomial distribution of a variable $X$ is specified by a set of *hyperparameters* $\{N'_x : x \in Val(X)\}$ as follows:

$$\Pr(\Theta_X) = \text{Dirichlet}(\{N'_x : x \in Val(X)\}) \propto \prod_x \theta_x^{N'_x - 1}.$$

Intuitively, the hyperparameters can be thought of as "pseudo counts", since they play a similar role to the actual counts we derive from the data. Under a Dirichlet prior, the probability of observing a sequence of values of $X$ with counts $N(x)$ is

$$\int \prod_x \theta_x^{N(x)} \Pr(\Theta_X \mid G) d\Theta_X =$$

$$\frac{\Gamma(\sum_x N'_x)}{\Gamma(\sum_x (N'_x + N(x)))} \times \prod_x \frac{\Gamma(N'_x + N(x))}{\Gamma(N'_x)},$$

where $\Gamma(x) = \int_0^\infty t^{x-1} e^{-t} dt$ is the *Gamma* function that satisfies the properties $\Gamma(1) = 1$ and $\Gamma(x+1) = x\Gamma(x)$.

Returning to the BDe score, let us assume that for each structure $G$, we have chosen the hyperparameters $N'^{(0)}_{i,j'_i,k'_i}$ and $N'^{\rightarrow}_{i,j_i,k_i}$. Then we can rewrite $\Pr(D \mid G)$ as a product of two terms,

$$\prod_i \prod_{j'_i} \frac{\Gamma(\sum_{k'_i} N'^{(0)}_{i,j'_i,k'_i})}{\Gamma(\sum_{k'_i} N'^{(0)}_{i,j'_i,k'_i} + N^{(0)}_{i,j'_i,k'_i})} \times$$

$$\prod_{k'_i} \frac{\Gamma(N'^{(0)}_{i,j'_i,k'_i} + N^{(0)}_{i,j'_i,k'_i})}{\Gamma(N'^{(0)}_{i,j'_i,k'_i})}$$

and similarly for the transition case.

This still requires us to supply the Dirichlet hyperparameters for each candidate DPN structure. Since the number of possible DPN structures is large, these prior estimates might be hard to asses in practice. Following [15], we can assign all of these given a *prior* DPN $B' = (B'_{(0)}, B'_\rightarrow)$ and two *equivalent sample sizes* $\hat{N}^{(0)}$ and $\hat{N}^\rightarrow$. Given these components, we assign the Dirichlet weights for $G = (G_0, G_\rightarrow)$ as follows:

$$N'^{(0)}_{i,j'_i,k'_i} = \hat{N}^{(0)} \times P_{B'_0}(X_i[0] = k'_i \mid \mathbf{Pa}(X_i[0]) = j'_i)$$

$$N'^{\rightarrow}_{i,j_i,k_i} = \hat{N}^\rightarrow \times P_{B'_\rightarrow}(X_i[1] = k_i \mid \mathbf{Pa}(X_i[1]) = j_i)$$

(Note that the choice of parents here is based on $G$, and might differ from the structure of $B'$.) Intuitively, we can consider the belief in $B'$ as equivalent to having previously experienced $\hat{N}^{(0)}$ sequences with $\hat{N}^\rightarrow$ transitions.

It is easy to verify that choosing priors in this manner preserves the main property we require: two DPN structures that imply identical independence assumptions will receive the same score. Thus, we claim that our definition is the natural extension of the BDe score to the case of dynamic probabilistic networks.

### 3.4  LEARNING IN PRACTICE

Both scores we considered so far have two important properties. First, the score of a DPN can be written as a sum of terms, (for the BDe case we look at the logarithm of $\Pr(G \mid D)$), where each term determines the score of a particular choice of parents for a particular variable. Thus, a local change to one family (e.g., arc addition or removal) affects only one of these terms. Second, the term that evaluates $X_i[t]$ given its parents is a function of the appropriate counts (either $N^{(0)}(\cdot)$ or $N^\rightarrow(\cdot)$) for that family. Thus, by caching these counts we can efficiently evaluate many families.

These two properties can be exploited by hill-climbing search procedures [3, 15] that gradually improve a candidate structure by applying the best arc addition, deletion, or reversal. In the case of DPNs, as opposed to static PNs, we have the additional constraint that the network structure must repeat over time. This reduces the number of options at each point in the search. Moreover, we can search for the best structure for $B_0$ independently of the search for the best structure for $B_\rightarrow$.

## 4  LEARNING FROM INCOMPLETE DATA

We now examine how to learn DPNs from incomplete data. Incomplete data is crucial since in most real life applications since we do not have complete observability of the process we want to model. This means that even if the process is a stationary Markov process, the partial observations we have are *not* Markovian. As an example, suppose we are tracking a car moving down the highway and we are observing the car's speed, lane, relative distance to other cars, and so on. This example is clearly not Markovian, given the observations. On the other hand, a Markovian model might be reasonable provided we include as part of the state information the driver's goals, e.g., get to the left lane, exit at the next off ramp, etc. By learning hidden variables, we can capture state information about the process. This allows the model to "remember" additional information about the past and to make better predictions of the future.

The main difficulty with learning from partial observations is that we no longer have the score decomposition properties of (4). This means that the optimal parameter choice in one part of the network depends on the parameter choices in other parts of the network. This problem can be understood better if we notice that once we have partial observability we can no longer talk about counts from the training data. For most events of interest, the counts are not defined, since we do not know the exact value of the variables in questions.

The most commonly used method to alleviate this problem is the *Expectation-Maximization* (EM) algorithm [7, 21]. The E-step of EM uses the currently estimated parameters to *complete* the data by computing the *expected* counts. The M-step then re-estimates the maximum-likelihood parameter values as if the expected counts were true observed counts. The central theorem underlying EM's behavior is that each EM cycle is guaranteed to improve the likelihood of the data given the model until it reaches a local maximum.

EM has been traditionally viewed as a method for adjusting the parameters of a fixed model structure. However, the underlying theorem can be generalized to apply to structural as well as parametric modifications. Friedman's *Structural EM* (SEM) algorithm [10] has the same E-step as EM, completing the data by computing expected counts based on the current structure and parameters. In addition to re-estimating parameters, the M-step of SEM can use the expected counts according to the current structure to evaluate any other candidate structure—essentially performing a complete-data structural search in the inner loop. Friedman shows that for a large family of scorings rules, including the BIC score and BDe score [11], the resulting network must have a higher score than the original. This is true even though the expected counts used in evaluating the new structure are computed using the old structure.

We can extend Friedman's results to the DPN case in the following way. Let $E[BIC((B_0', B_\rightarrow') : D^+) : D, (B_0, B_\rightarrow)]$ be the expected BIC score of a DPN $(B_0', B_\rightarrow')$ based on all possible completions $D^+$ of the data, i.e., an assignment of values to the hidden variables. The expectation is taken with respect to $\Pr(D^+ \mid D, (B_{(0)}, B_\rightarrow))$, i.e., the probability assigned to this completion based on the old DPN. Using Theorem 3.1 of [10], we can prove the following:

**Theorem 4.1:**

$$BIC((B_0', B_\rightarrow') : D) - BIC((B_0, B_\rightarrow) : D) >$$
$$E[BIC((B_0', B_\rightarrow') : D^+) : D, (B_0, B_\rightarrow)] -$$
$$E[BIC((B_0, B_\rightarrow) : D^+) : D, (B_0, B_\rightarrow)]$$

That is, if we choose a new DPN that has a higher expected score than the expected score of the old DPN, then the *true* score of new DPN will also be higher than the *true* score of the old DPN. Moreover, the difference in the expected scores is a lower bound on the improvement in terms of the actual score we are trying to optimize.

The crucial property of the expected BIC score is that it, too, decomposes into a sum of local terms, as follows. By the linearity of expectation, we can "push" the $E$ operator through the summation signs implicit in (6) to get an equation which involves terms of the following form:

$$\begin{aligned} E[\vec{N}_{i,j_i,k_i}] &= \sum_\ell \sum_t E[I(X_i[t] = k_i, \mathbf{Pa}(X_i[t]) = j_i; \mathbf{x}^\ell)] \\ &= \sum_\ell \sum_t \Pr(X_i[t] = k_i, \mathbf{Pa}(X_i[t]) = j_i | \mathbf{x}^\ell) \end{aligned}$$

and similarly for $E[N^{(0)}_{i,j_i',k_i'}]$. These are called the *expected sufficient statistics*. Hence the key requirement to apply the SEM algorithm in the incomplete data case is the ability to compute the probabilities of all the families $\Pr(X_i[t] = k_i, \mathbf{Pa}(X_i[t]) = j_i | \mathbf{x}^\ell)$ of all the networks which we wish to evaluate, i.e., the current network and the "neighboring" networks in the search space.

To efficiently compute the probabilities of the families, we can convert the DPN to a join tree and use a two-pass dynamic programming algorithm [19], similar to the forwards-backwards algorithm used in HMMs [29]. To efficiently compute the probability of a set of nodes that is not contained in any of the join tree nodes, we need to use more sophisticated techniques [30]. A simpler approach, which is the one we currently use, is to connect together all the variables in a single timeslice into a single node (i.e., to convert the DPN to a Markov chain), compute the expected number of transitions between every pair of consecutive states in this chain, and then marginalize these counts to get the counts for each family. We also maintain a cache of the expected counts computed so far based on the current completion model, and use this cache to avoid recomputing expected counts. Whatever technique we use, computing exact expected sufficient statistics for all models of interest is likely to remain computationally challenging. In the discussion section we mention some approximation techniques that may allow us to learn larger models.

In summary, the SEM procedure is as follows.

procedure DPN-SEM:
  Choose $(B_0^0, B_\rightarrow^0)$ (possibly randomly).
  Loop for $n = 0, 1, \ldots$ until convergence
    Improve the parameters of $(B_0^n, B_\rightarrow^n)$ using EM
    Search over DPN structures
      (using expected counts computed with $(B_0^n, B_\rightarrow^n)$)
    Let the best scoring DPN seen be $(B_0^{n+1}, B_\rightarrow^{n+1})$
    if $(B_0^{n+1}, B_\rightarrow^{n+1}) = (B_0^n, B_\rightarrow^n)$
      return $(B_0^{n+1}, B_\rightarrow^{n+1})$

| Data Size | | BIC | | | BDe | | |
|---|---|---|---|---|---|---|---|
| $N_{seq}$ | $N_{\rightarrow}$ | 0 | 1 | 2 | 0 | 1 | 2 |
| 250 | 11387 | -1.4323 | -1.4196 | -1.0812 | -1.5286 | -1.5027 | -1.3347 |
| 500 | 28043 | -1.0737 | -1.0248 | -1.0168 | -1.2503 | -1.0160 | -1.0339 |
| 1000 | 61864 | -0.9979 | -0.9961 | -0.9783 | -0.9638 | -0.9647 | -0.9676 |
| 1500 | 102906 | -0.9820 | -0.9773 | -0.9787 | -0.9625 | -0.9511 | -0.9888 |

Figure 2: Logloss (bits) per slice on test data for the networks learned in the driving domain using the BIC and BDe scores. Columns are labeled by the number of additional binary variables introduced.

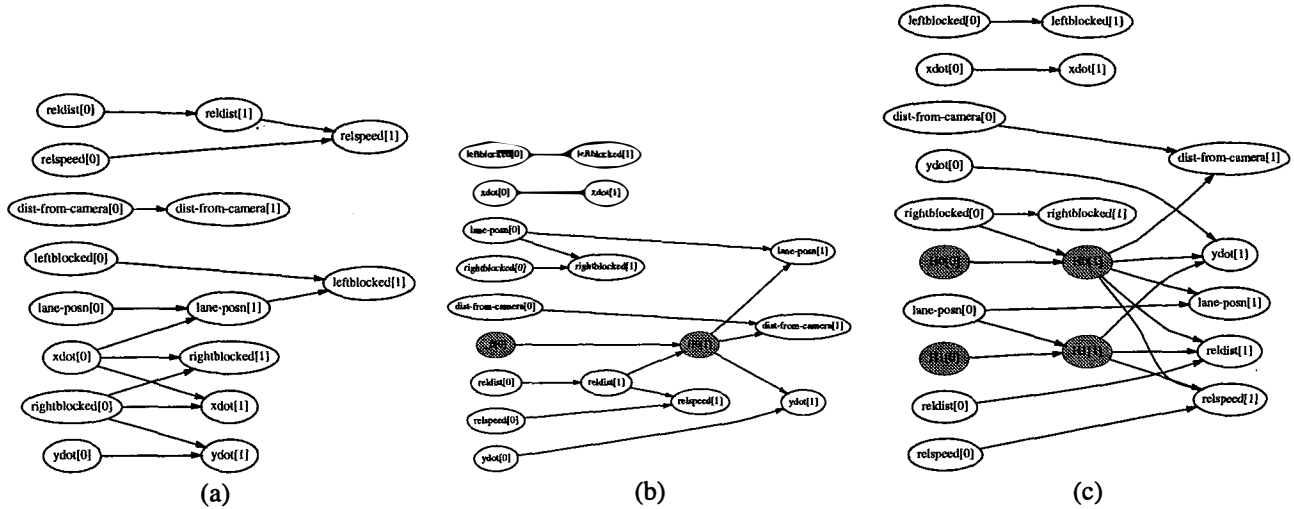

Figure 3: Transition models learned from the driving domain. Shaded nodes correspond to hidden variables.

The above discussion was for the application of SEM for the BIC score. In [11], Friedman shows how to extend the SEM procedure to learn with the BDe score. The details are more involved, since the BDe score is not linear. Nonetheless, Friedman shows that it is a reasonable approximation to use the BDe score on the expected counts inside the SEM loop.

## 5 APPLICATIONS

In this section we describe two preliminary investigations that attempt to evaluate the usefulness of the DPN technology we develop here for real-life applications.

### 5.1 INFERRING DRIVER BEHAVIOR

In many tracking domains, we would like to learn a predictive model of the behavior of the object being tracked. The more accurate the model, the more robust the tracking system and the more useful its predictions will be in decision making. For objects with hidden state, it is our hope that DPNs can be learned that correctly reflect the unobserved process governing the behavior. The learned model may also provide insight into how the behavior is generated.

In this section, we describe some experiments we carried out using a simulated driving domain [9]. The data is an idealization of what cameras mounted on the side of the road can collect. In particular, at each time step of the simulation, we get a report on cars that are within the camera's range. The report for each car has the following attributes: position and velocity (relative to the camera's reference frame), relative speed and distance to the car in front, and whether there is a car immediately to the left or right. From this data, we want to learn models of typical classes of driving behavior. Such models can be useful for several tasks. A prime example arises in the BATmobile autonomous car project [8]. The BATmobile's autonomous controller attempts to predict the behavior of neighboring car. For example, it would be useful to know that someone who has just driven across two lanes might be attempting to leave the freeway, and consequently is likely to cut in front of you. Since tracking information from real cameras is readily available [24], it is reasonable to hope that realistic models of human drivers can be obtained. In addition to their use in autonomous vehicles, such models are of paramount importance in so-called "microscopic" traffic models used in freeway design and construction planning and also in safety studies.

In our experiments, we generate a variety of simulated traffic patterns consisting of populations of vehicles with a mixture of different driving tendencies (e.g., trucks, sports cars, Sunday drivers, etc.). We generated 3500 cars and tracked their behavior over sequences of roughly 40-70 time steps. The observed data were discretized using fixed-sized bins. We then trained networks using a dataset consisting of the first 250, 500, 1000, and 1500 sequences, and tested these networks on the last 2000 sequences.

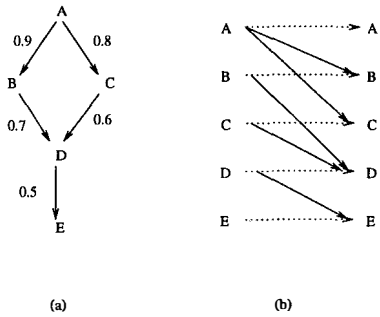

Figure 4: (a) A simple pathway model with five vertices. Each vertex represents a site in the genome, and each arc is a possible triggering pathway. (b) A DPN model that is equivalent to the pathway model shown in (a).

We learned networks with 0, 1, and 2 hidden variables using decision tree CPTs [12] with the BIC and BDe score. The initial network structure in each case had each hidden variable (if any) dependent on its value in the previous time slice, and each observable variable dependent only on the hidden variables in the same time slice. (Thus, we did not put in any persistence arcs in the initial model.) This structure connects each hidden variable to all the other variables, allowing that variable to carry forward information about previous time slices.

Table 2 summarizes the logloss, per time slice, on the test data for the different networks we learned. We can roughly interpret the negative of the logloss as the number of bits needed to encode a time slice using each network.[3] These numbers indicate that the addition of hidden variables improves the predictive ability of the models. They also indicate that we might benefit from using many more training sequences; this is because many of the important events such as lane changes are quite rare in typical freeway traffic patterns. Some of the learned networks are shown in Figures 3. The networks seem to include at least some of the essential relationships that we expect to see. For example, in Figure 3(b), the single hidden node has parent "relative distance" and children "lane-position" and "longitudinal velocity" (but not "lateral velocity"). This suggests that it encodes "need to take avoiding action."

## 5.2 LEARNING CAUSAL PATHWAYS IN BIOLOGICAL PROCESSES

DPNs are a powerful representation language for describing causal models of stochastic processes. Such models are useful in many areas of science, including molecular biology, where one is often interested in inferring the structure of regulatory pathways. McAdams and Shapiro [26] model part of the genetic circuit of the lambda bacteriophage in terms of a sequential logic circuit; and attempts have even be made to automatically infer the form of such Boolean circuits from data [22]. However, it is well known that the abstraction of binary-valued signals together with deterministic switches often breaks down, and one must model the continuous nature and inherent noise in the underlying system to get accurate predictions [25]. Also, one must be able to deal with noisy and incomplete observations of the system.

---
[3] For comparison, the gzip utility compresses the observations to approximately 5.6 bits per time slice.

We believe that DPNs provide a good tool for modeling such noisy, causal systems, and furthermore that SEM provides a good way to learn these models automatically from noisy, incomplete data. Here we describe some initial experiments using DPNs to learn small artificial examples typical of the causal processes involved in genetic regulation. We generate data from models of known structure, learn DPN models from the data in a variety of settings, and compare these with the original models. The main purpose of these experiments is to understand how well DPNs can represent such processes, how many observations are required, and what sorts of observations are most useful. We refrain from describing any *particular* biological process, since we do not yet have sufficient real data on the processes we are studying to learn a scientifically useful model.

Simple genetic systems are commonly described by a *pathway model*—a graph in which vertices represent genes (or larger chromosomal regions) and arcs represent causal pathways (Figure 4(a)). A vertex can either be "off/normal" (state 0) or "on/abnormal" (state 1). The system starts in a state which is all 0s, and vertices can "spontaneously" turn on (due to unmodelled external causes) with some probability per unit time. Once a vertex is turned on, it stays on, but may trigger other neighboring vertices to turn on as well—again, with a certain probability per unit time. The arcs on the graph are usually annotated with the "half-life" parameter of the triggering process. Note that pathway models, unlike PNs, can contain directed cycles. For many important biological processes, the structure and parameters of this graph are completely unknown; their discovery would constitute a major advance in scientific understanding.

Pathway models have a very natural representation as DPNs: each vertex becomes a state variable, and the triggering arcs are represented as links in the transition network of the DPN. The tendency of a vertex to stay "on" once triggered is represented by persistence links in the DPN. Figure 4(b) shows a DPN representation of the five-vertex pathway model in Figure 4(a). The nature of the problem suggests that noisy-ORs (or noisy-ANDs) should provide a parsimonious representation of the conditional density function at each node. To specify a noisy-OR for a node with $k$ parents, we use parameters $q_1, \ldots, q_k$, where $q_i$ is the probability the child node will be in state 0 if the $i$th parent is in state 1. (Thus we need only $k$ parameters instead of $2^k$ for a full CPT.) In the five-vertex DPN model that we used in the experiments reported below, all the $q$ parameters (except for the persistence arcs) have value 0.2. For a strict persistence model (vertices stay on once triggered), $q$ parameters for persistence arcs are fixed at 0. To learn such noisy-OR distributions, we follow the technique suggested in [27]; this entails introducing a new hidden node for each arc in the network, which is a noisy version of its parent, and replacing each noisy-OR gate with a deterministic-OR gate. We also tried using gradient descent, following [1], but encountered difficulties with convergence in cases where the optimal parameter values were close to the boundaries (0

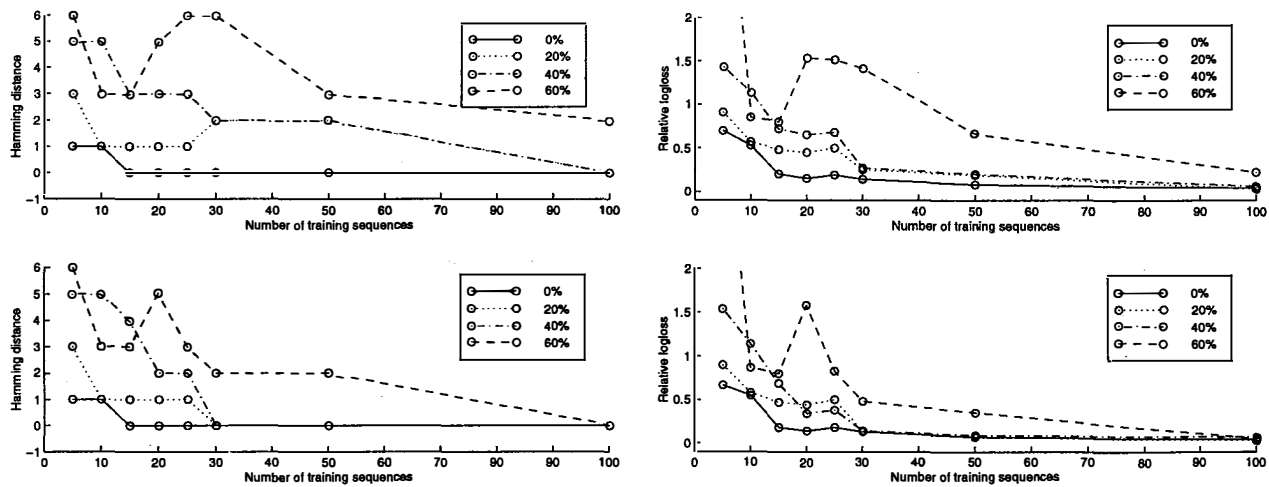

Figure 5: Hamming distance (left) of the learned models compared to the generating model for the five-vertex process, and relative log-loss (right) compared with the generating model on independent sample of 100 sequences. Tests are for regimes in which 0%, 20%, 40%, and 60% of the slices are hidden at random. Top: tabulated CPTs; bottom: noisy-OR network.

or 1).

In all our experiments, we enforced the presence of the persistence arcs in the network structure. We used two alternative initial topologies: one that has only persistence arcs (so the system must learn to add arcs) and one that is fully interconnected (so the system must learn to delete arcs). Performance in the two cases was very similar.

We experimented with three observation regimes that correspond to realistic settings:

- The complete state of the system is observed at every time step.
- Entire time slices are hidden uniformly at random with probability $h$, corresponding to intermittent observation.
- Only two observations are made, one before the process begins and another at some unknown time $t_{obs}$ after the process is initiated by some external or spontaneous event. This might be the case with some disease processes, where the DNA of a diseased cell can be observed but the elapsed time since the disease process began is not known. (The "initial" observation is of the DNA of some other, healthy cell from the same individual.)

This last case, which obtains in many realistic situations, raises a new challenge for machine learning. We resolve it as follows: we supply the network with the "diseased" observation at time slice $T$, where $T$ is with high probability larger than the actual elapsed time $t_{obs}$ since the process began.[4] We also augment the DPN model with a hidden "switch" variable $S$ that is initially off, but can come on spontaneously. When the switch is off, the system evolves according to its normal transition model $P(\mathbf{X}[t+1] \mid \mathbf{X}[t], S = 0)$, which is to be determined from the data. Once the switch turns on, however, the state of the system is frozen—that is, the conditional probability distribution $P(\mathbf{X}[t+1] \mid \mathbf{X}[t], S = 1)$ is fixed so that $\mathbf{X}[t+1] = \mathbf{X}[t]$ with probability 1. The persistence parameter for $S$ determines a probability distribution over $t_{obs}$; by fixing this parameter such that (a priori) $t_{obs} < T$ with high probability, we effectively fix a scale for time, which would otherwise be arbitrary. The learned network will, nonetheless, imply a more constrained distribution for $t_{obs}$ given a pair of observations.

---

[4] With the $q$ parameters set to 0.2 in the true network, the actual system state is all-1s with high probability after about $T = 20$, so this is the length used in training and testing.

We consider two measures of performance. One is the number of different edges in the learned network compared to the generating network, i.e., the Hamming distance between their adjacency matrices. The second is the difference in the logloss of the learned model compared to the generating model, measured on an independent test set. Our results for the five-vertex model of Figure 4(a) are shown in Figure 5. We see that noisy-ORs perform much better than tabular CPTs when the amount of missing data is high. Even with 40% missing slices, the exact structure is learned from only 30 examples by the noisy-OR network. However, when all-but-two slices are hidden, the system cannot learn effectively with a reasonable amount of data (results not shown). The case in which we do not even know the time at which the second observation is made (which we modeled with the switching variable) is even harder to learn (results not shown). Obviously, the application of prior knowledge will be very important in reducing the data requirements.

## 6 DISCUSSION

In this paper we addressed the question of learning the structure and parameters of DPNs from complete and incomplete data. To the best of our knowledge, we are the first to examine this problem. As our experiments show, we can learn non-trivial structures from synthetic data and "realistic" data from a nontrivial simulator.

The main bottleneck in the application of our procedures is inference, which is necessary to compute the expected sufficient statistics. Unlike the case of PNs, a sparse DPN structure does not necessarily ensure fast inference—the minimum size of the posterior distribution for a slice is generally exponential in the number of variables that have parents in the previous slice. There are various approximations that we could use to speed up inference: (1) The method proposed by Boyen and Koller [2], which approximates posterior probabilities in the DPN in a factored form; this should be particularly appropriate for the biological models we are investigating. (2) Stochastic simulation—for example, the ER/SOF algorithm of Kanazawa et al. [18]. (3) Variational approximations, e.g., Jaakkola and Jorden [17] and Ghahramani and Jordan [14]. (4) Methods based on multilevel abstraction hierarchy to detect which variables are related to each other.

We are currently extending SEM to learn the structure of linear Gaussian DPNs; we hope this will prove competitive with traditional techniques of system identification [23]. One advantage of the Gaussian case over the discrete case is that marginalizing the posterior over two slices is an efficient operation. Ultimately we wish to tackle the case of hybrid DPNs, with both discrete and continuous variables. The advantage of hybrid DPNs over switching state space models [16] is that the state variables can be represented in factored form. For example, in the driving domain, we can have separate variables for the continuous observations (such as speed and position) and for the discrete hidden states (such as "want to change lane" or "want to overtake"). The question of how to know when to add hidden variables is a very interesting one which we are also currently investigating.

### Acknowledgments

We thank Jeff Forbes and Nikunj Oza for their help in getting the training data for the driving domain. This work was supported in part by ARO under the MURI program "Integrated Approach to Intelligent Systems", grant number DAAH04-96-1-0341.